\documentclass[preprint,12pt,3p]{elsarticle}

\usepackage{geometry}
\usepackage{placeins}

\usepackage{caption}
\usepackage{graphicx}
\usepackage{url}
\usepackage{subfigure}
\usepackage[hidelinks]{hyperref}
\usepackage{booktabs}
\usepackage{threeparttable}
\usepackage{multirow}
\usepackage{float}
\usepackage{hyperref}
\usepackage{cleveref}
\usepackage{longtable}

\usepackage{graphicx}
\usepackage{amssymb}

\usepackage{lineno}

\usepackage{xcolor}
\usepackage{ulem}

\definecolor{nyupurple}{RGB}{150, 35, 140}
\newcommand{\xzhangdelete}[1]{\textcolor{nyupurple}{\sout{#1}}}

\journal{arXiv}

\begin{document}

\begin{frontmatter}


\title{A Clustering-aided Ensemble Method for Predicting Ridesourcing Demand in Chicago}



\author[label1]{Xiaojian Zhang}
\address[label1]{Department of Civil and Coastal Engineering, University of Florida}
\ead{xiaojianzhang@ufl.edu}

\author[label1]{Xilei Zhao\corref{cor1}}
\ead{xilei.zhao@essie.ufl.edu}

\cortext[cor1]{Corresponding author. Postal address: 1949 Stadium Rd, Gainesville, FL 32611, USA.}

\begin{abstract}
Accurately forecasting ridesourcing demand is important for effective transportation planning and policy-making. With the rise of Artificial Intelligence (AI), researchers have started to utilize machine learning models to forecast travel demand, which, in many cases, can produce higher prediction accuracy than statistical models. However, most existing machine-learning studies used a global model to predict the demand and ignored the influence of spatial heterogeneity (i.e., the spatial variations in the impacts of explanatory variables). Spatial heterogeneity can drive the parameter estimations varying over space; failing to consider the spatial variations may limit the model’s prediction performance. To account for spatial heterogeneity, this study proposes a Clustering-aided Ensemble Method (CEM) to forecast the zone-to-zone (census-tract-to-census-tract) travel demand for ridesourcing services. Specifically, we develop a clustering framework to split the origin-destination pairs into different clusters and ensemble the cluster-specific machine learning models for prediction. We implement and test the proposed methodology by using the ridesourcing-trip data in Chicago. The results show that, with a more transparent and flexible model structure, the CEM significantly improves the prediction accuracy than the benchmark models (i.e., global machine-learning and statistical models directly trained on all observations). This study offers transportation researchers and practitioners a new methodology of travel demand forecasting, especially for new travel modes like ridesourcing and micromobility.

\end{abstract}

\begin{keyword}
Ridesourcing demand \sep Spatial heterogeneity \sep Clustering \sep Machine learning \sep Ensemble model


\end{keyword}

\end{frontmatter}



\section{Introduction}
\label{S:1}

As one of the pioneers in the shared economy, ridesourcing providers, such as Uber, Lyft and Didi Chuxing, have experienced exponential growth in recent years and significantly impacted people's way of life \citep{wang2019ridesourcing}. Although the rapid development of ridesourcing services has brought travelers tremendous convenience, the proliferation of this new mobility option is also considered as a disruptive force, which may bring negatively impacts to the environment and social equity \citep{henao2019impact, clewlow2017disruptive, brown2018ridehail}. Therefore, to maximize the benefits and alleviate the adverse effects brought by this emerging travel mode, transportation planners and engineers should spend efforts to accurately model and forecast the ridesourcing demand. To this end, more accurate and easy-to-implement measures should be conducted to help further understand and predict the ridesourcing demand.

Traditional travel demand measure is the four-step model, including trip generation, trip distribution, mode split, and traffic assignment. An alternative to the four-step model is the direct demand model \citep{talvitie1973direct}. The direct demand model can be used for modeling Origin-Destination-pair (OD-pair) demand. It combines the first three steps in the four-step model so that it is easy to implement and can avoid the accumulated errors produced by each step \citep{choi2012analysis}. As plenty of OD-pair data are gradually available, more and more studies have applied the direct demand model to forecast the travel demand \citep[e.g.][]{choi2012analysis, ding2019does, yan2020using, shao2020threshold}. Traditionally, the direct demand model refers to a statistical approach, which takes a predetermined (i.e., multiplicative) model form \citep{talvitie1973direct}. Even though the statistical models are easy to implement and interpret, they usually suffer a low prediction accuracy due to the fixed model structure that violates the nonlinear nature of the real-world data. Recently, multiple studies have applied machine learning models using the direct demand modeling framework, and have verified the their predictive strength \citep{ding2019does, yan2020using, shao2020threshold, gan2020examining, xu2021identifying, geng2019spatiotemporal}. Machine learning models are usually non-parametric; and they can effectively identify the nonlinear relationships between the input and the target variable. These merits provide machine learning models more flexibility, resulting in a superior prediction performance. Accordingly, the evidence suggests that machine learning is a promising application in direct demand modeling to further improve the travel demand prediction.

Although the emerging machine learning models are proved to have strong predictive capability in travel demand forecasting, two imperative issues that are highly related to the prediction accuracy still remain unsolved. First, some current studies only tested the global models (i.e., models that fit on the entire observations). They treated all observations as a whole while ignoring that travel demand has a location-specific nature and is sensitive to urban forms \citep{chen2019discovering, yu2019exploring, qian2015spatial, ghaffar2020modeling}. Functionalities of urban forms at different locations can strongly shape the transportation needs, resulting in travel demand varying over space \citep{qian2015spatial}. The global models are not able to provide local estimations, which may restrict the prediction accuracy and bias the interpretations. Second, machine learning models generally follow the assumption of variable stationarity over space and rarely take spatial heterogeneity of built environment into account \citep{chen2021nonlinear}. In fact, the influence of built environment characteristics on travel demand has strong spatial variations \citep{chen2019discovering, qian2015spatial,ma2018geographically, ding2021non}. Failing to account for these inherent spatial variabilities may limit machine learning models' capability to capture the interactions between independent variables, resulting in biased predictions. In addition, the potential spatial heterogeneity may hide certain characteristics that are locally associated with travel demand, causing machine learning modeling results less reliable. Therefore, to address the spatial heterogeneity and further improve the prediction accuracy, we should consider the context-specific information in machine learning models while improving the models' capabilities to cope with spatial variations.

To address these issues, we proposed a machine-learning-based ensemble planning framework called Clustering-aided Ensemble Method (CEM). In CEM, we (1) adopt knowledge-driven and data-driven clustering methods (such as $K$-Means) to partition the data samples (OD-pair data); (2) use machine learning models (e.g., random forest, gradient boost decision tree, support vector machine and neural network) to build cluster-specific submodels; and (3) combine all submodels to form an ensemble learning model to make cluster-specific predictions.

The first step for CEM is to partition all samples into some subgroups through clustering methods. The reason why we choose clustering here is 2-fold. Firstly, clustering is a widely accepted technique in spatial classification and pattern recognition \citep{mennis2009spatial, jain2000statistical}, which is appropriate to identify the underlying OD-pair subgroups. Secondly, clustering methods, such as $K$-Means and Latent Class Cluster Analysis, have been extensively applied in analyzing transportation data and have shown their superior capabilities in addressing heterogeneity \citep[e.g.][]{varone2018understanding, soria2020k, depaire2008traffic, chang2019investigating}. Thus, we propose to use clustering methods to segment the OD pairs to accommodate spatial heterogeneity. The clustering mechanism can be based on domain knowledge and the data. Specifically, when purely domain-knowledge-based clustering (i.e., manually segmenting the data based on domain knowledge) is applied, the results may seem explainable and easy to understand; however, we cannot ensure the heterogeneity of each cluster through this approach \citep{depaire2008traffic}. On the other hand, even if numerous data-driven clustering methods have shown the strength in providing high-quality and insightful results, in some circumstances, the clustering results may be less reliable and hard-to-explain when data have high dimensionalities \citep{dasgupta2020explainable, holzinger2018machine}. We decide to combine domain knowledge and data-driven clustering to account for this trade-off, i.e., we use both knowledge-driven clustering and data-driven clustering. Some recent studies have found that the injection of prior domain knowledge can boost the performance of data-driven clustering \citep{forestier2010collaborative,pedrycz2004fuzzy}. Therefore, we first manually segment some OD pairs based on domain knowledge and then apply data-driven clustering methods over the remaining OD pairs.

The second foremost step of CEM is to build the cluster-specific submodels and combine them for predictions. Recently some researchers have explored that combining clustering and machine learning models into ensemble can help machine learning models achieve higher prediction accuracy \citep{trivedi2011clustering, trivedi2015utility, mueller2019cluster, jurek2014clustering}. They suggested that by partitioning the whole dataset into several clusters, the ensemble can model the interrelationships between variables more precisely from the cluster-specific perspective and improve the bias-variance trade-off while accommodating more variations. Therefore, we hypothesize that as the unobserved context-specific information is accommodated in modeling, the ensemble is expected to achieve better prediction accuracy. Although some researchers in other fields have applied similar ideas to tackle their problems of interest, no priors work has yet developed or implemented such a method in travel demand forecasting.

In view of the aforementioned discussions, we use the Chicago ridesourcing-trip data as a case study to build the CEM as well as the benchmark models (directly trained on the whole observations). Empirical results show that compared to the benchmark model, our proposed ensemble-learning framework has considerably improved the model performance metrics, i.e., the accuracy improvement rates examined by the mean absolution error (MAE) and root mean square error (RMSE) is 14.36\% and 6.78\%, respectively. In addition to improving the prediction accuracy, this study will also guide researchers, planners or policymakers in exploring key determinants and providing valuable insights associated with ridesourcing demand from the cluster-specific perspective. 

The main contributions of this study are summarized as follows:
\begin{itemize}
    \item This paper applies knowledge-driven and data-driven clustering to identify the underlying heterogeneous OD-pair groups of ridesourcing demand in the City of Chicago.
    \item This paper presents a new method, called Clustering-aided Ensemble Method (CEM), which integrates clustering and supervised learning techniques for ridesourcing demand modeling. The CEM can effectively address the spatial heterogeneity and significantly outperform the benchmark models in forecasting ridesourcing demand.

\end{itemize}

The remainder of this study is as follows. Section \ref{LR} summarizes the existing literature related to this study. In Section \ref{Method}, we describe the overall methodological framework, and the clustering and machine learning methods used in this paper. Section \ref{S:4} introduces the dataset used for analysis. Section \ref{S:Results} showcases the results. Section \ref{S:Conclusion} concludes our paper by summarizing key findings, proposing valuable discussions and suggesting limitations with future research directions.

\section{Literature Review}
\label{LR}

How to accurately model the travel demand has been dominating transportation research for decades. Scholars have studied the travel demand modeling under various contexts, for example, the ridesourcing demand \citep[e.g.][]{yan2020using,xu2021identifying,saadi2017investigation, yu2019exploring}, transit ridership \citep[e.g.][]{ding2019does, gan2020examining, cervero2010direct, ma2018geographically}, taxi demand \citep[e.g.][]{qian2015spatial} and bikesharing system \citep[e.g.][]{chen2016dynamic, li2015traffic}. In the past several years, both traditional statistical and machine-learning methods were widely applied to model the travel demand. These studies provide valuable insights and decision-making suggestions for transportation professionals.

\subsection{Statistical Methods in Travel Demand Modeling}

Previous studies often apply traditional statistical methods to model the travel demand, and explore its relationship to the potentially contributing factors. Multiple Ordinary Least Square (OLS) regression is one of the most popular models being used to model the travel demand, due to its simplicity and interpretability.  For example, \citet{zhao2013influences} used Multiple OLS regression to model the metro station ridership in Nanjing, China. \citet{cervero2010direct} used multiple OLS regression to predict the bus rapid transit patronage in Southern California. Since the travel demand is usually considered to be count data, many studies have adopted the Poisson Regression model and the multiplicative model. To name a few, \citet{choi2012analysis} investigated the relationship between the built environment and station-to-station metro ridership, using both the Poisson Regression model and the multiplicative model. \citet{marquet2020spatial} applied a Poisson model to evaluate the associations between the ridesourcing demand and neighborhood characteristics. These models are computationally tractable and easy to interpret. But they are usually global models and cannot account for spatial heterogeneity, thus having limited predictive power \citep{chen2019discovering}. In fact, the impacts of the explanatory variables such as built environment, are proved to be heterogeneous over space \citep{tu2018spatial, ma2018geographically, yu2019exploring}. The heterogeneous spatial effects, determined by urban forms, have strong associations with the travel demand \citep{chen2019discovering, qian2015spatial, ghaffar2020modeling}. Failing to account for the spatial heterogeneity may bring bias to the parameter estimation and lead to erroneous modeling results.

To tackle this, some researchers used geographically weighted model or generalized additive models \citep{yu2019exploring, ma2018geographically, ghaffar2020modeling, lavieri2018model}. For example, \citet{yu2019exploring} applied Geographically Weighted Poisson Regressions (GWPR) to investigate the spatial relationship between built environment and weekdays and weekend ridesourcing demand. Results show that GWPR has \textit{R}-square over 0.9, indicating a superior goodness-of-fit. Using the ridesourcing-trip data from the City of Chicago, \citet{ghaffar2020modeling} proposed to employ a random-effects negative binomial (RENB) regression approach to identify the associations between ridesroucing-trip usage and the built-environment variables. By adding a random term to represent the location-specific effects, RENB effectively captures the variations over space. These models can capture the location-specific effects and then estimate the parameters locally, thus enjoying superior prediction accuracy. Also, the model results offer us a deeper understanding of how spatial heterogeneity shapes travel demand in geographical context \citep{chen2019discovering}.

Nevertheless, one major issue of the statistical regression models is that they are restricted to the predefined model structure (e.g., linear or log-linear) and only allow specific assumptions for parameters and error term distributions. This limitation may result in inaccurate parameter estimations and thus bias model interpretations. Hence, more flexible models should be taken into account, if we attempt to improve the accuracy of travel demand modeling.

\subsection{Machine Learning in Travel Demand Modeling}

Recently, machine learning has gained increasing popularity, and has been widely accepted in tackling transportation problems. Due to its flexible model structure and strong capabilities in dealing with empirical data, some prior studies have adopted machine learning methods to model the nonlinear relationships between key factors and travel demand, e.g., \citep{yan2020using, ding2019does, gan2020examining, xu2021identifying}. Specifically, \citet{ding2019does} adopted gradient boost decision trees (GBDT) to model the relationship between built-environment characteristics and metrorail ridership. \citet{gan2020examining} applied the GBDT model to investigate the nonlinear associations between built environment and station-to-station ridership. \citet{xu2021identifying} applied the random forest model to explore the key factors associated with the ridesplitting adoption rate. \citet{yan2020using} used an ensemble learning model (i.e., random forest) in ridesourcing direct demand modeling. These studies collectively suggest that machine learning approaches can effectively model the underlying nonlinearity and provide an accurate estimate of travel demand. In addition, some recent publications also evaluated deep learning models when forecasting travel demand and verified their exceptional predictive strength \citep[e.g.][]{geng2019spatiotemporal,ke2018hexagon, li2020forecasting}. For example, \citet{ke2018hexagon} developed three hexagon-based convolutional neural networks to forecast ridesourcing demand. \citet{liu2019attention} developed an attention-based deep ensemble net to forecast the taxi-hailing demand and validated its exceptional predicting capabilities through real-world datasets.

Collectively, these studies motivate us that machine learning models can be powerful tools for travel demand prediction. However, we will not examine complex deep learning  models (except for a simple neural network model) in this study, due to the following \xzhangdelete{three} reasons. First, due to the complex nature of the networks along with the requirement of large-scale training samples, fitting the deep learning  models usually consumes excessive computational resources \citep{8622396}. Second, complex deep learning  models usually serve for dynamic or real-time demand predictions, which may not be applicable in this study as we prefer to predict the ridesouricng demand during an extended period.  Hence, complex deep learning  models can hardly fit into the scope of this study.

Even though machine learning has a great potential in improving travel demand predictions, most of them are utilized under a global analytical context. Currently, few studies have addressed spatial heterogeneity in machine learning models. Recently, scholars have evaluated integrating the idea of GWR into random forest, i.e., constructing local decision trees for each spatial unit, to model the transit ridership \citep{chen2021nonlinear}. Results show that by incorporating spatial heterogeneity into machine learning, the resulting model can boost the prediction capability than the traditional machine learning models. This result inspires us to consider the heterogeneous spatial effects when modeling travel demand.

\subsection{Clustering-aided Ensemble Learning}

Clustering is a powerful machine learning technique that has been widely applied in travel behavior analysis. Recently, clustering has gained extensive popularity in solving spatial classification and pattern recognition problems to promote travel behavior modeling. Specifically, a series of studies have applied clustering to extract underlying groups that share distinct context-specific travel behavior \citep[e.g.,][]{varone2018understanding,soria2020k, jia2019hierarchical, shanqi90understanding}. For example, \citet{soria2020k} used \textit{$K$-Prototypes}, an unsupervised-learning clustering technique, to detect underlying heterogeneous ridesourcing-service user segments. They identified six unique user groups and discussed their distinct travel behavior. \citet{shanqi90understanding} applied $K$-Means to identify groups of vulnerable populations that possess heterogeneous travel behavior based on a set of indicators of participation in bus activities. Results illustrated that the travel behavior of different population groups has strong spatial variations. These studies suggested the promising potential of leveraging clustering to identify the spatial variations across the City of Chicago.

In recent years, there is a trend of integrating clustering and machine learning techniques to build ensembles (i.e., the aggregation of a group of models) to improve prediction accuracy. This approach usually consists of two separate steps: (1) create a group of independent cluster-level submodels; (2) combine them to form an ensemble \citep{trivedi2011clustering}. Due to its flexibility, this framework has been applied in various fields. To name only a few, \citet{mueller2019cluster} introduced an ensemble-learning framework with the incorporation of clustering to help evaluate rates of health insurance coverage in Missouri. They found that with the incorporation of clustering as a preprocessing technique, the proposed framework outperforms independent prediction models and generates valuable cluster-specific interpretations. \citet{jurek2014clustering} used Cluster-Based Classifier Ensemble (CBCE) for activity recognition within smart environments, and results showed that CBCE achieved higher accuracy than the single classifiers.
Nevertheless, to the best of our knowledge, no previous research has applied similar methodological frameworks to OD-pair travel demand modeling (especially for ridesourcing trips). Following the discussions above, it is promising to integrate clustering with advanced supervised-learning methods to create ensembles. Since spatial variations are captured and modeled by clustering and cluster-specific machine learning submodels respectively, the ensemble is expected to achieve higher prediction accuracy.

\section{Methodology}
\label{Method}

\subsection{Methodological Framework}
\label{Section 3.1}
\begin{figure}[!ht]
  \centering
  \includegraphics[width=1\textwidth]{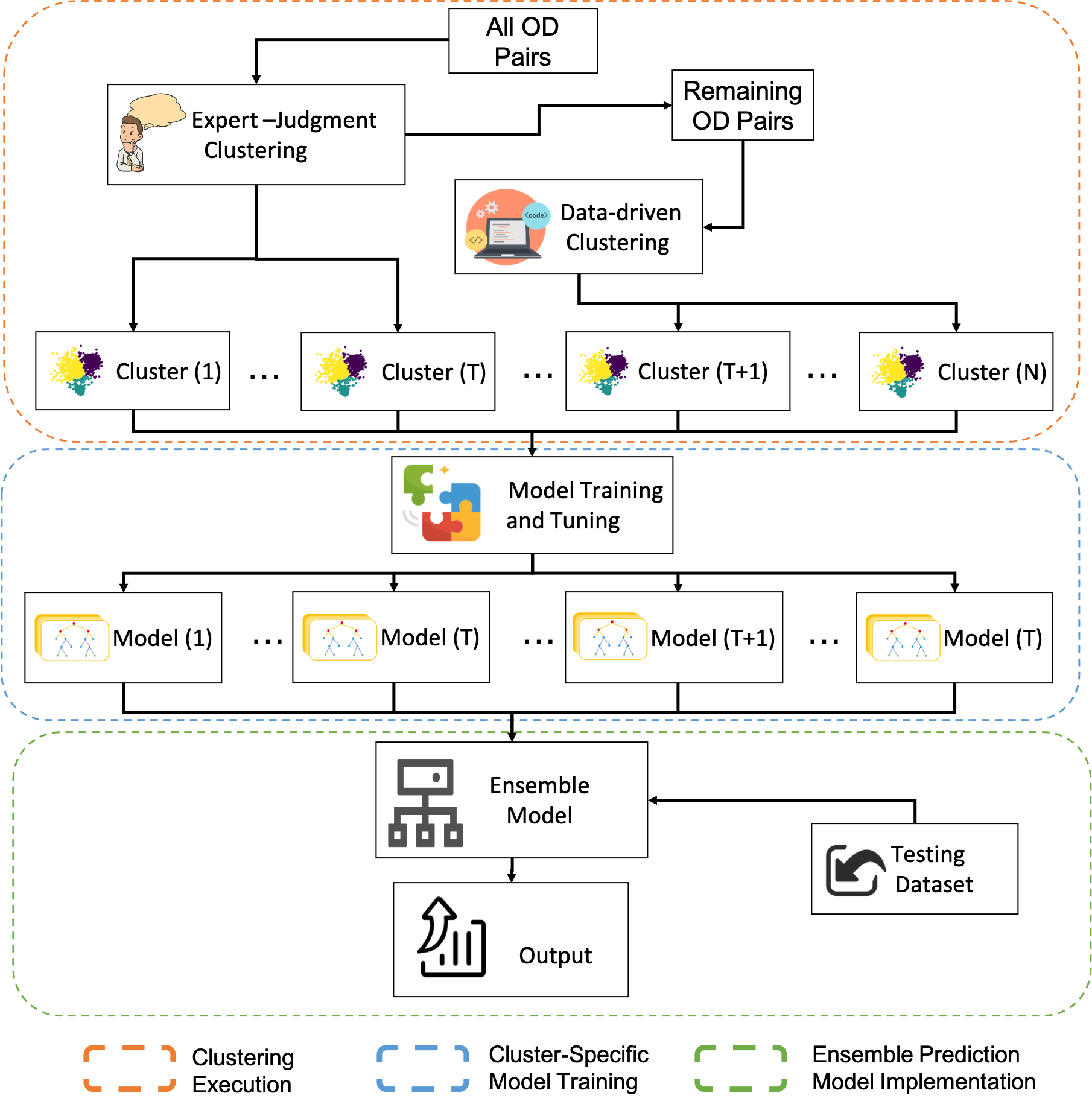}
  \caption{Framework of the Clustering-aided Ensemble Method}\label{fig:workflow}
\end{figure}

Figure \ref{fig:workflow} provides a detailed diagram outlining the methodological framework of CEM. The procedure contains three major steps, including \textit{Clustering Execution}, \textit{Cluster-Specific Model Training}, and \textit{Ensemble Model Implementation}.

\textit{Clustering Execution} contains two main steps, namely, a knowledge-driven clustering followed by a data-driven clustering. Knowledge-driven clustering is used to manually separate \textit{T} special clusters from the whole dataset. These clusters' characteristics or patterns are usually context-specific and hard-to-classify; and we are especially interested in these special clusters. If we blindly include them into data-driven clustering, these special clusters may not be fully separated. Thus, we first use the domain knowledge to create such clusters manually, and then apply data-driven clustering to the rest of the data.

\textit{Cluster-Specific Model Training} includes two main steps: base-learner training and tuning. In the base-learner training process, we will train cluster-specific machine learning models. Although various machine learning models can be applied here, four tree-based models, Support Vector Machine and Neural Network are selected in this study. This is because these machine learning models have been widely used in ridesourcing demand modeling \citep[e.g.,][]{saadi2017investigation, yan2020using} and display an outstanding model performance. Also, these models can cope with high-dimensional input variables in noisy environments and produces high-level prediction accuracy. In the base-learner tuning step, \textit{Grid Search} and \textit{Cross Validation} (CV) are used to find the optimal hyperparameters.

\textit{Ensemble Model Implementation} involves two steps: (1) aggregating all base learners to constitute the ensemble model, and (2) using the cluster-specific submodels to complete the prediction tasks.

\subsection{Knowledge-driven Clustering}

With the development of machine learning models, there is a growing trend to make machine-learning results more interpretable and explainable \citep{roscher2020explainable}. Even if pure data-driven clustering may achieve an better result, some special contexts (e.g., lacking good-quality data) may impact its reliability of being an ideal fit \citep{borghesi2020improving}. In this study, if we use all input variables (including travel impedance) into data-driven clustering, we may obtain a hard-to-explain result that ignores the context-specific characteristics of different regions. In the meantime, the unreliable clustering result may bring great challenges to cluster-level submodels' development. Since the goal of clustering is to identify the heterogeneous clusters while exploring in-depth context-specific insights, it is of great importance to improve the interpretability of clustering results.  \citet{borghesi2020improving} indicated that integrating domain knowledge\footnote{In \citep{von2019informed}, \textit{Knowledge} is defined as ``validated information about relations between entities in certain contexts''.} into machine learning models can help promote the model's performance, lower the model's complexity, and simplify the data-training process. Also, \citet{von2019informed} pointed out that one commonly-used method to integrate domain knowledge into machine learning models is preprocessing the training data, e.g., feature engineering or removing outliers. Accordingly, driven by the knowledge related to ridesourcing trips, we will first separate some special OD pairs from the original dataset before data-driven clustering.

\subsection{Data-driven Clustering}
Clustering is usually defined as dividing a large dataset into several groups by minimizing intra-cluster similarity and maximizing inter-cluster differences. Conventionally, data-driven clustering methods can be categorized into two major classes: supervised and unsupervised clustering. Supervised clustering methods require a labeled training data sample to map the labels for new testing samples, while unsupervised clustering methods direct at exploring undetected patterns in a dataset without pre-existing labels \citep{russell2002artificial, hinton1999unsupervised}. Since there are no previously known OD-pair categories available in this study, unsupervised clustering is the preferred option. 

In general, unsupervised methods can be classified into two groups: partitioning and hierarchical clustering \citep{sarle1990algorithms, arabie1996hierarchical}. While hierarchical clustering is less sensitive for initialization and can be easily perceived by people, it suffers from its high computational cost (high time and space complexity), so it is rarely used in large-scale clustering. Conversely, partitioning clustering methods have high efficiency in dealing with large-scale datasets and have linear time complexity \citep{madhulatha2012overview}. Additionally, partitioning methods are more popularly accepted in pattern recognition than hierarchical methods \citep{jain2000statistical}. Since we have a large-scale dataset containing a great number of variables in this study, partitioning clustering is selected for achieving the clustering task.

\textit{K-Means} is one of the most widely used partitioning algorithms, it groups \textit{N} observations into \textit{K} clusters, minimizing a criterion known as \textit{inertia} or within-cluster variances \citep{madhulatha2012overview}. 

Also, \textit{Gaussian-Mixture Model} (GMM) plays an important role in partitioning clustering, it employs Expectation-Maximization algorithm, which is usually used for parameter estimation, to divide the dataset into \textit{K} components, and each component is based on Gaussian probability density function \citep{reynolds2009gaussian}.

The result of \textit{K-Means} is that each observation is assigned to one of the clusters, while Gaussian-Mixture Model gives the probability that these data points are assigned to each cluster. Both \textit{K-Means} and \textit{Gaussian-Mixture Model} are suitable for mining large-scale datasets and have a computationally fast speed. Therefore, they are chosen for clustering in our study. Nonetheless, one of the drawbacks of these two methods is that it is challenging to preset the optimal number of clusters (components). In our study, repeated iterations are performed to find the optimal result to tackle this. Additionally, Davies Bouldin Index (DBI) \citep{davies1979cluster}, which evaluates intra-cluster similarity and inter-cluster differences, is applied to help decide the optimal number of clusters. The smaller the value of DBI, the better the clustering results.

\subsection{Prediction Models}
Machine learning models were broadly applied in travel demand modeling in previous research \citep[e.g.][]{yan2020using, ding2019does, ke2018hexagon, shao2020threshold}. Among these, the tree-based model plays an important role. Tree-based models refer to machine learning methods with a tree-like model structure and employ the Decision Tree (DT) as the fundamental model. Tree-based models, such as RF and GBDT, can flexibly and efficiently accommodate various types of input variables (e.g., missing values, outliers, categorical or continuous variables) \citep{breiman2001random}. Additionally, tree-based models do not have the predetermined linear model structure like statistical models. Additionally, they can model the non-linear interrelationships between high-dimensional input variables and the target variable (i.e., ridesourcing demand)\citep{breiman2001random}. Furthermore, tree-based models are insensitive to the multicollinearity issues that often exist in the conventional regression models. These merits allow us to predict the ridesourcing demand accurately and explore how various types of variables shape the ridesourcing demand. In addition to the tree-structured models, Support Vector Machine and Neural Network model are also being widely used due to their flexibility and superior predictive strengths. Hence, these two models may also be an appropriate fit for this study.  

In a nutshell, we examined four commonly-used tree-based models (i.e., DT, RF, GBDT and XGBoost), along with Support Vector Machine and Neural Network in forecasting ridesourcing demand. We also fit several traditional regression models for comparison. Now we briefly discuss these models. 

\subsubsection{Decision Tree}

Decision Tree (DT) is the fundamental model of Classification and Regression Tree (CART) \citep{breiman1984classification}. Conceptually, an impurity measure should be predefined to evaluate how impure a node is. DT makes the regression by starting from the root and split it iteratively, each time makes the node more impure. In DT, each internal node of the tree makes the split based on the value of a single feature and the terminal nodes represent the predicted value of the target variable. DT models are simple to understand and interpret. However, they are susceptible to overfit the training sample \citep{hastie2009elements}.

\subsubsection{Random Forest}
To avoid the overfitting issues of general CART models, the ensemble methods were proposed to form more robust, stable and accurate models instead of a single DT \citep{breiman1996bagging}. Random Forest is a widely-used ensemble method. Generally, RF is a collection of several DTs, each of which is slightly different from the other \citep{breiman2001random}. Even though DT is sensitive to overfit the training sample, by averaging the predictions of these trees, RF has the ability to yield accurate predictions while remaining a low risk of overfitting. Firstly, based on the bootstrapping rule, a random subset of the training sample is selected with replacement for every single tree. Secondly, a "feature bagging", which means selecting a subset of the features at each splitting node, is used to construct a single DT. Then all DTs constitute an RF; and these DTs collectively determine the prediction results. For regression purposes, all results are averaged to form a final prediction, instead of "soft voting" \citep{zhou2012ensemble}, which are applied in classification problems. 

\subsubsection{Boosting}
Boosting is another well-known ensemble method \citep{friedman2001greedy}. Boosting method builds the trees (week learners) in sequential, and they gradually reduce the bias of the previous trees. The final predictions are obtained by combining the weighted results of each boosted weak learner. The higher accuracy the learner achieves, the more weight it gets assigned. In this paper, we applied two tree-based boosting models: Gradient Boosting Decision Tree (GBDT) and eXtreme Gradient Boosting (XGBoost).

GBDT develops a sequence of successive DTs with each added DT trained to fit the residual errors\footnote{Here, the squared-error \(L_2\) loss function is tested.} (the difference between the observed value and the predicted value) made by the previous tree. Specifically, the added DTs are used to reduce the residual errors to make the loss decrease follow the negative gradient direction in each iteration, i.e., to minimize the difference between the observed value and the predicted value in the fastest direction. The final prediction is calculated based on the sum of results of all DTs. GBDT is often considered to have a better predictive capability than traditional regression models and some machine-learning methods (e.g., support vector machine and neural networks \citep{ding2018applying}). 

Based on the framework of GBDT, XGBoost was introduced by \citep{chen2016xgboost}. XGBoost is now receiving continuously increased attention due to its strong and robust predictive performance. Compared with GBDT, XGBoost has two main changes: 1) XGBoost introduces the regularization term in the loss function in order to lower the variance and address overfitting issues; and 2) instead of only using the first-order derivative like GBDT, XGBoost employs the second-order Taylor series (including both first and second-order derivative) in the loss function. 

\subsubsection{Support Vector Machine}
The Support Vector Machine (SVM) model was first developed as a classification tool \citep{cortes1995support}, and has been extended to many regression studies in recent years. SVM regression model aims to find the optimal hyperplane that minimizes the sum of the distance from the data points to the hyperplane. SVM usually has a good performance when fitted with specified (linear or nonlinear) kernel. However, SVM is susceptible to overfitting problems when selecting the nonlinear kernel, even if the nonlinear kernel usually has a better performance on real-world data \citep{cawley2010over}. 

\subsubsection{Neural Network}
A basic Neural Network (NN) model architecture contains three layers: input layer with each node representing a variable, hidden layer with each node representing the intermediate variables and output layer with the final calculations. In this study, we use a single-hidden-layer NN for prediction. Specifically, the data will first be fed into the input layer, and then be linearly calculated at the hidden layer based on the variance and bias, and finally end up at the output layer, being nonlinearly computed with activation function. 

\subsubsection{Traditional Regression Model}
In this study, we also fit a linear regression model, a log-transformed linear regression model and a Poisson regression model for comparison, since they are frequently used in travel demand modeling \citep[e.g.,][]{choi2012analysis,zhao2013influences, marquet2020spatial}. These models have a predefined model structure and assume the independent variables have a linear or log-linear relationships with the outcome variable. This assumption may mask the real relationships and cause the parameter estimations biased. However, these models can be directly interpreted from the coefficients and provide level of significance for each coefficient estimate \citep{xu2021identifying, ghaffar2020modeling, marquet2020spatial}.

\subsection{Hyperparameter Optimization}
\label{hyperparameters optimization}
Almost every machine learning model has the specific hyperparameters that need to be tuned. Likewise, hyperparameters should be exclusively adjusted based on the dataset. Instead of setting the hyperparameters manually, we adopted \textit{Grid Search} (i.e., an exhaustive search that tries every specific value of each hyperparameter to fit the model) to find the optimal value of each hyperparameter. The calibration of the hyperparameters for each model is presented in Appendix.A.

\subsection{Model Comparison}

We use a 5-fold \textit{Cross Validation} (CV) to examine the model's predictive performance. The first goal of CV is testing the model's ability to predict new data that are not included in training. Another one is to avoid overfitting and selection bias problems. For the 5-fold CV, the entire dataset is first randomly divided into five subsets. Then while holding out one subset for testing, we train the model using the remaining four subsets. After training, we use the model to make predictions for the hold-out subset. The predicted values are compared with the true values to evaluate the model's out-of-sample predictive ability. We repeat this process for each subsets, totally five times, and average the results of each model to get a final mean estimate of the predictive performance.

There are several performance metrics for measuring the predictive capabilities of a model. For this study, we select \textit{Mean Absolute Error} (MAE), \textit{Mean Square Error} (MSE) and \textit{Root Mean Square Error} (RMSE) as metrics. We use MSE as the criterion for hyperparameters optimization and cluster-specific model selection, while MAE and RMSE for examining the model's predictive capabilities of the testing set. Specifically, MAE, MSE and RMSE are defined as follows:

\begin{equation}
    MAE = \frac{1}{N}\sum_{k=1}^{N}\left|\widehat{y_k}-y_k\right|,
    \label{Eq: MAE}
\end{equation}

\begin{equation}
    MSE = \frac{1}{N}\sum_{k=1}^{N}(\widehat{y_k}-y_k)^2,
    \label{Eq: MSE}
\end{equation}

\begin{equation}
    RMSE = \sqrt{\frac{1}{N}\sum_{k=1}^{N}(\widehat{y_k}-y_k)^2},
    \label{Eq: RMSE}
\end{equation}

\noindent where \textit{N} is the number of observations of the training sample, \(y_k\) represents the \(k^{th}\) observed (true) value of the independent variable, \(\widehat{y_k}\) is the \(k^{th}\) predicted value of the independent variable. 

While both MAE, MSE and RMSE have been commonly used by various studies to assess a model's performance, there is still no consensus on the most appropriate metric. While MAE gives the same weight to all errors, MSE/RMSE is more sensitive to variance as it gives errors with larger absolute values more weight than errors with smaller absolute values, which means MSE/RMSE is more susceptible to outliers \citep{chai2014root}. After each model is tuned, we compute the model's out-of-sample (through testing dataset) MAE and RMSE to evaluate the predictive capability.

\section{Data}
\label{S:4}
Chicago ridesourcing-trip data being analyzed in this paper are publicly available at Chicago Data Portal\footnote{\url{https://data.cityofchicago.org/Transportation/Transportation-Network-Providers-Trips/m6dm-c72p/data}}. This study collected data from November 1, 2018 to March 31, 2019, including 45,338,599 trips. Every observation in the raw data has plenty of attributes, but only the following are considered: fare, trip miles, trip seconds (duration), pick-up/drop-off location. To avoid privacy issues, the City of Chicago has rounded the fare and time to the nearest \$2.50 and 15 minutes, respectively. Also, the pick-up/drop-off locations were aggregated at the census tract level.

To prepare the data for clustering and modeling, we processed the data in the following procedures. First, due to the protection of passengers' privacy\footnote{\url{http://dev.cityofchicago.org/open\%20data/data\%20portal/2019/04/12/tnp-taxi-privacy.html}}, a significant proportion of trips reported only has the community ID and the geographic coordinates of the pick-up/drop-off community areas' centroid instead of the tract ID and its coordinates. We proposed a method called \textit{Stratified Assignment}\footnote{Based on the observed probability distribution, we randomly assigned the pick-up/drop-off tract ID to the trips only having community ID for origin/destination. For a detailed description of the inference procedure, please see \citep{xu2021identifying}.} to infer the tract ID for these trips. Second, we removed all trips that occurred on weekends and federal holidays. Furthermore, we aggregated the data at the OD-pair level and removed the outliers \footnote{We first removed trips with trip fare equal to 0, trip duration less than 1 minute and trip distance less than 0.25 miles. We also considered that the distance and duration should be relatively close for trips sharing the same OD pair. Thus we defined that trips whose distance or duration was more than three interquartile ranges away from either the upper or the lower quartile were outliers. }. After the outlier removal, we computed the following seven travel-related attributes: total number of trips, median of trip fare, distance and duration, standard deviation of trip fare, distance and duration. We also calculated the \textit{Euclidean Distance} between each two census tracts' centroids. Then, we dropped the OD pairs with less than 50 trips to avoid the impact of randomness on demand modeling. After the procedure above, the total number of OD pairs considered in this study was 67,498, including 752 census tracts as trip destinations and 755 census tracts as trip origins.

To accurately model the ridesourcing demand in Chicago, we further included three categories of independent variables: socio-economic and demographic factors, built-environment and transit supply characteristics. Socio-economic and demographic data were collected from the American Community Survey 2013–2017 5-year estimates data; some population, employment and worker-related characteristic were obtained from the 2015 Longitudinal Employer-Household Dynamics data, and the crime rate data were downloaded from Chicago Data Portal. In addition, we supplemented the dataset with some transit-supply characteristics through General Transit Feed Specification (GTFS) data, and adopted geographical information system (GIS) tools to calculate some built-environment variables. We also obtained the work score of each census tract's centroid from \href{http://walkscore.com}{WalkScore.com} API. All the abovementioned variables were aggregated at the census tract level. \autoref{tab:Variable Description} provides a detailed variable description. 

Since the scales of the input variables vary broadly and their units are different, the objective function in some machine learning algorithms will not work properly. For example, in KMeans algorithm, the similarity of two points is measured by their Euclidean distance. If one of the features has a wide range of values, the outcome will be significantly controlled by this feature and calculated with a bias. Transforming the data into comparable scales can effectively address this issue. Thus, we applied \textit{Min-Max Normalization} to scale the range of the values of each data point into [0,1] to make sure each variable is equally measured in clustering. Nevertheless, we used the pre-normalized data in descriptive statistics (see in Appendix.C) to make it more straightforward for readers to understand.

\begin{table}[!ht]
\caption{Variable Description}
\label{tab:Variable Description}
\begin{threeparttable}
\resizebox{\textwidth}{!}{%
\begin{tabular}{@{}ll@{}}
\toprule
Variable                                & Description                                        \\ \midrule
\textbf{Dependent Variable}                       &
                            \\
Total\_number\_trips                    & Total number of ridesourcing trips        \\
\textbf{Independent Variable}                       &
                            \\
\textit{Travel-impedance Variables}               &                                                    \\
Fare\_median                                      & Median of trip fare (\$)
\\
Fare\_sd                                          & The standard deviation of trip fare (\$)           \\
Miles\_median                                     & Median trip distance (mile)                        \\
Miles\_sd                                         & The standard deviation of trip distance (mile)     \\
Seconds\_median                                   & Median trip duration (second)                      \\
Seconds\_sd                                       & The standard deviation of trip duration (second)   \\

\textit{Socio-economic and Demographic Variables}$^a$ &                                                    \\
Pcttransit                                        & Percentage of workers taking transit to work       \\
Pctmidinc        & Percentage of middle-income households (\$50 k to \$75 k)                          \\
Pctmale                                           & Percentage of male population                      \\
Pctsinfam                                         & Percentage of single-family homes                  \\
Pctmodinc        & Percentage of moderate-income households (\$25 k to \$50 k)                        \\
Pctyoung                                          & Percentage of population aged 18–44                \\
Pctwhite                                          & Percentage of White population                     \\
Pcthisp                                           & Percentage of Hispanic population                  \\
Pctcarown                                         & Percentage of households with at least one car     \\
Pctrentocc                                        & Percentage of renter-occupied housing units        \\
Pctasian                                          & Percentage of Asian population                     \\
Pctlowinc                                         & Percentage of low-income households (\$25 k less)  \\
CrimeDen                                          & Density of violent crime                           \\
PctWacWorker54$^b$   & Percentage of workers (workplace) workers aged 54 or younger                     \\
PctWacLowMidWage$^b$ & Percentage of workers (workplace) with earnings \$3333/month or less             \\
PctWacBachelor$^b$   & Percentage of workers (workplace) with bachelor's degree and above               \\
Commuters        & Total number of commuters (from origin census tract to destination census tract) \\
\textit{Built-environment Variables}              &                                                    \\
Popden                                            & Population density                                 \\
IntersDen                                         & Intersection density                               \\
EmpDen                                            & Employment density                                 \\
EmpRetailDen                                      & Retail employment density                          \\
Walkscore                                         & Walkscore of centroid of census tract              \\
RdNetwkDen                                        & Road Network Density                               \\
\textit{Transit-supply Variables}                 &                                                    \\
SerHourBusRoutes                                  & Aggregate service hours for bus routes             \\
SerHourRailRoutes                                 & Aggregate service hours for rail routes            \\
PctBusBuf                                         & Percentage of tract within 1/4 mile of a bus stop  \\
PctRailBuf                                        & Percentage of tract within 1/4 mile of a rail stop \\
BusStopDen                                        & Number of bus stops per square mile                \\
RailStationDen                                    & Number of rail stops per square mile               \\ \bottomrule
\end{tabular}%
}
\flushleft{\footnotesize{$^a$Variables listed above are at the tract-to-tract level, the others are at the tract level.}}
\flushleft{\footnotesize{$^b$Variable codes with "Wac" represents the characteristics of workers aggregated at the area of their workplace.}}

\end{threeparttable}
\end{table}

\section{Results}
\label{S:Results}
This section presents the final results of every best cluster-specific model with optimal hyperparameters, knowledge-driven and data-driven clustering, and prediction accuracy comparison between CEM and the benchmark models.

\subsection{Clustering Results}


\subsubsection{Knowledge-driven Clustering}
Previous research showed that the airport should be considered as a special external zone in trip-distribution model \citep{lavieri2018model}, and airport-related trips were separated as a small segment through clustering due to its distinctive characteristics \citep{soria2020k}. According to these findings, we explore the relevant characteristics of two airport-related census tracts (O'Hare International Airport and Midway International Airport) in the . Compared with the characteristics of the other census tracts, theirs are significantly different. These pieces of evidence encourage us to model the airport-related OD pairs separately. Similarly, trips that take place at downtown were also proved to showcase a special pattern. For example, \citet{shokoohyar2020determinants} found that in Philadelphia, the average waiting time of Uber in city center was the lowest among all other areas, indicating that Uber is more accessible around the center city. In addition, the population density, employment density and the accessibility to transit are extremely high in downtown areas in Chicago, these variables have been shown a positive relationship with ridesourcing demand \citep{yan2020using, yu2019exploring, ghaffar2020modeling,marquet2020spatial}. These empirical results confirm the need to separate downtown-related OD pairs from the entire dataset before data-driven clustering.

\begin{figure}[!ht]
\centering
\includegraphics[width=1\linewidth]{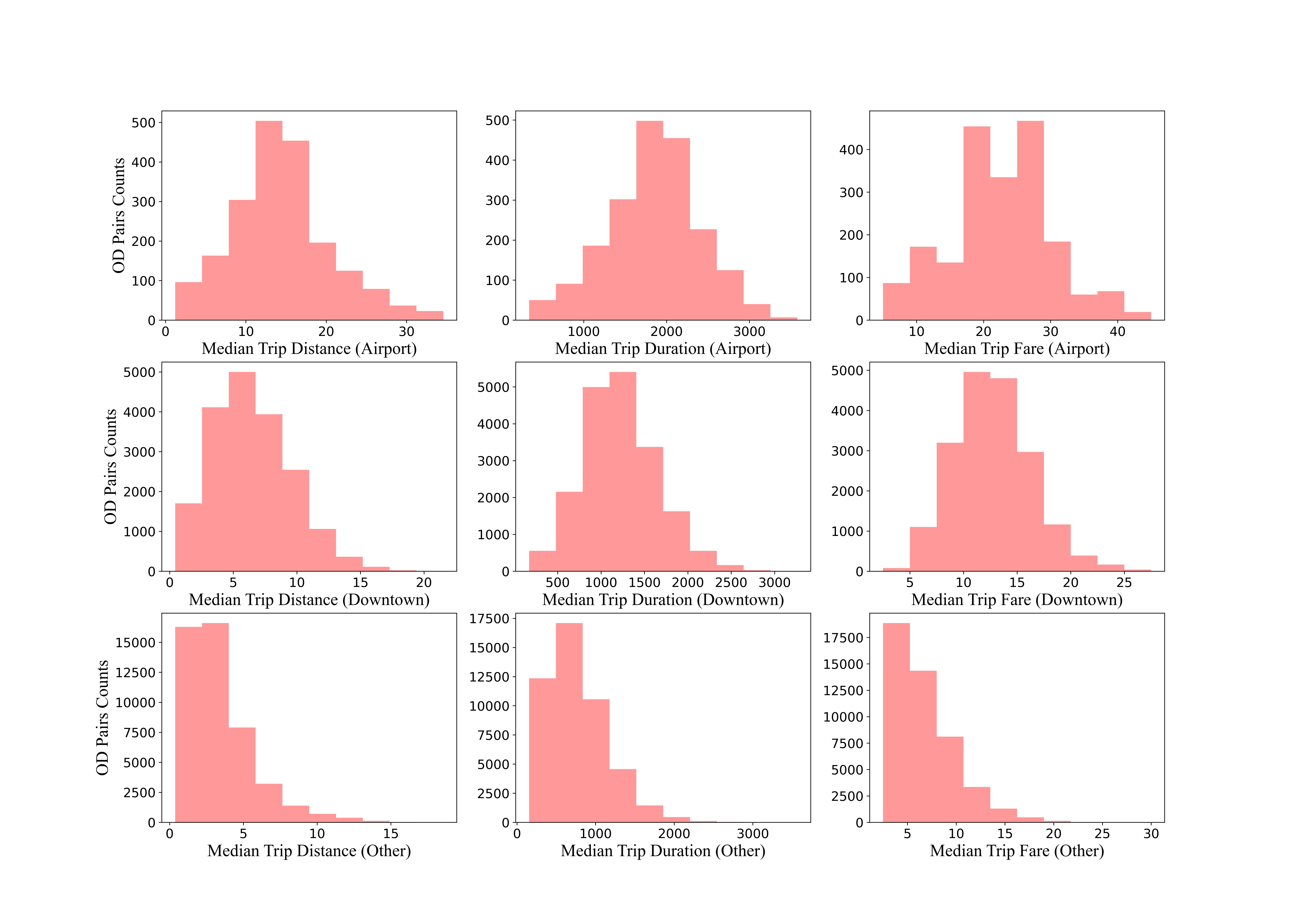}
\caption{Histograms of key trip characteristics of Airport/Downtown-related OD pairs and the remaining OD pairs}

\label{Fig: Pattern Difference}
\end{figure}

Apart from the hints from previous studies, we also perform an exploratory analysis. \autoref{Fig: Pattern Difference} presents the histogram of trip fare, trip distance and trip duration of the airport and downtown-related OD pairs. The distribution of three key travel-impedance characteristics of Airport and Downtown related OD Pairs are different. As \autoref{Fig: Pattern Difference} shows, other (the remaining) OD pairs' median-trip-distance distribution was close to an exponential shape, while airport-related and downtown-related OD pairs' median-trip-distance distribution displayed a normal-like shape. Distributions of median trip duration and median trip fare also show this similar pattern, indicating that the characteristics of airport and downtown-related trips are special.

Hence, in this study, we first split the OD pairs containing airport ($n$ = 1,981) and downtown ($n$ = 18,868) tracts from the dataset as two special clusters.

\subsubsection{Data-Driven Clustering}

Dependent variable and travel-impedance variables were excluded in data-driven clustering, because (1) including travel impedance variables in the clustering step may create endogenous issues (i.e., the other three types of variables may shape travel impedance); and (2) one recent studies showed that categorizing the city into several clusters based on the non-travel variables offered us a clear view of the city structure, which promotes our investigation of the travel behavior across the whole city \citep{shokoohyar2020determinants}.

During the estimation phase, we applied $K$-Means and GMM on the remaining data with DBI as the performance metric. The models were developed by setting the number of clusters to range from 2 to 7 with one hundred different random seeds. Experiments showed that $K$-Means outperformed GMM and 3 was the optimal number of clusters. The detailed result of the data-driven clustering is shown in Appendix.B.

\autoref{tab:Descriptive Statistics} (in Appendix.C) summarizes in detail the data-driven clustering result along with the knowledge-driven clustering result and the descriptive statistics of each cluster. Note that: (1) variable codes with ``\_Ori'' refers to variables measured at trip origin; (2) we exclude variables code with ``\_Des'' from this table because the values of these characteristics at trip origins and destinations are quite similar and also due to the limited space. (3) variable codes with ``\_HW'' or ``\_WH'' represents the number of commuters from home to work or from work to home, respectively; (4) to make it more transparent for readers, we present the descriptive statistics by using the pre-normalized data.

\subsection{Model Selection}
We develop nine different models for five cluster-specific models and the benchmark model with the entire training set. We first randomly hold out 90\% of the data for training and the remaining 10\% for testing, and then tune each model using \textit{Grid Search} with a 5-fold CV. Several key hyperparameters of different models are examined, and \autoref{tab:Optimal Values of Hyperparameterse} in the appendix presents the optimal value of each hyperparameter tuned. Based on the fine-tuned models, we perform the prediction and calculate the MAE and RMSE for the testing set. In hyperparameter tuning and model selection part, we uniformly use MSE as the performance metrics. 

\autoref{tab:Submodel Selection} presents the optimal cluster-specific submodel of each cluster. The GBDT outperforms the others among four cluster-specific subsets except for cluster 3, where XGBoost has the best performance. For the benchmark model (fit on the whole training set), we also select the GBDT.

From \autoref{tab:Submodel Selection}, RF and two boosting models outperform a single DT in cluster-level prediction. This finding further confirms the previous studies that DT is more susceptible to noise and vulnerable to overfitting issues \citep{last2002improving}. In addition, in most cases, traditional statistical models have worse predictive performance than machine learning models. This is unsurprising because the traditional regression models usually have a predetermined model structure so that they have limited flexibility to model the nonlinearity. Finally, it is somewhat surprising that DT, SVM and NN even underperform the traditional regression models (e.g., linear regression, Poisson regression) in Cluster 1. One possible explanation is that the Cluster 1, with only 3\% observations, is insufficient for machine learning models to fully study, while may be enough for the traditional statistical models.




\begin{table}[!ht]
\centering
\caption{Cluster-specific Model Selection}
\label{tab:Submodel Selection}
\resizebox{0.8\textwidth}{!}{%
\begin{tabular}{@{}ccccccc@{}}
\toprule
Models & Cluster 1 & Cluster 2 & Cluster 3 & Cluster 4 & Cluster 5 & All Clusters \\ \midrule
DT & 6419900.8 & 920547.6 & 13723.1 & 15310.0 & 30712.6 & 708823.0 \\
RF & 613028.9 & 492186.4 & 8458.5 & 12335.1 & 16687.6 & 132548.0 \\
GBDT & \textbf{335159.9} & \textbf{253110.9} & 6182.7 & \textbf{8603.3} & \textbf{14262.1} & \textbf{101220.6} \\
XGBoost & 386634.3 & 282288.5 & \textbf{6001.4} & 9628.0 & 14848.8 & 111163.2 \\
SVM & 1672082.7 & 2288033.5 & 13939.9 & 19406.6 & 45412.7 & 740750.3 \\
ANN & 1828347.1 & 1049879.9 & 12465.5 & 15326.0 & 25990.5 & 372186.1 \\
LR & 937607.7 & 1886376.0 & 12481.5 & 14993.2 & 27205.7 & 736534.0 \\
Log-LR & 2942888.3 & 3719873.0 & 17451.1 & 15877.5 & 43840.7 & 884279.8 \\
Poisson & 1110374.0 & 1108225.0 & 12977.8 & 14263.1 & 28692.6 & 518824.1 \\ \bottomrule
\end{tabular}%
}
\parbox[t]{0.78\textwidth}{\vskip3pt{\footnotesize Note: The lowest MSE values for each cluster are emphasized in \textbf{bold}. Their corresponding submodels will be used in CEM.}}
\end{table}


\begin{table}[!ht]
\centering
\caption{Cluster-Specific Proportion (\%) of OD Pairs and Average Ridesourcing Demand}
\label{tab:OD_Pair_Characteristics}
\resizebox{\textwidth}{!}{%
\begin{tabular}{ccccc}
\hline
No. & Cluster & Number of OD Pairs & Share (\%) & Average Ridesourcing Demand \\ \hline
1 & Airport Cluster & 1981 & 2.93 & 819.30 \\
2 & Downtown Cluster & 18868 & 27.95 & 646.00 \\
3 & Low-Income Cluster & 12523 & 18.55 & 135.87 \\
4 & Moderate-Income Cluster & 11717 & 17.36 & 128.50 \\
5 & High-Income Cluster & 22409 & 33.20 & 195.74 \\
- & Original Dataset & 67498 & 100.00 & 317.12 \\ \hline
\end{tabular}%
}
\end{table}

\begin{figure}[!htbp]
  \centering
  \includegraphics[scale = 0.53]{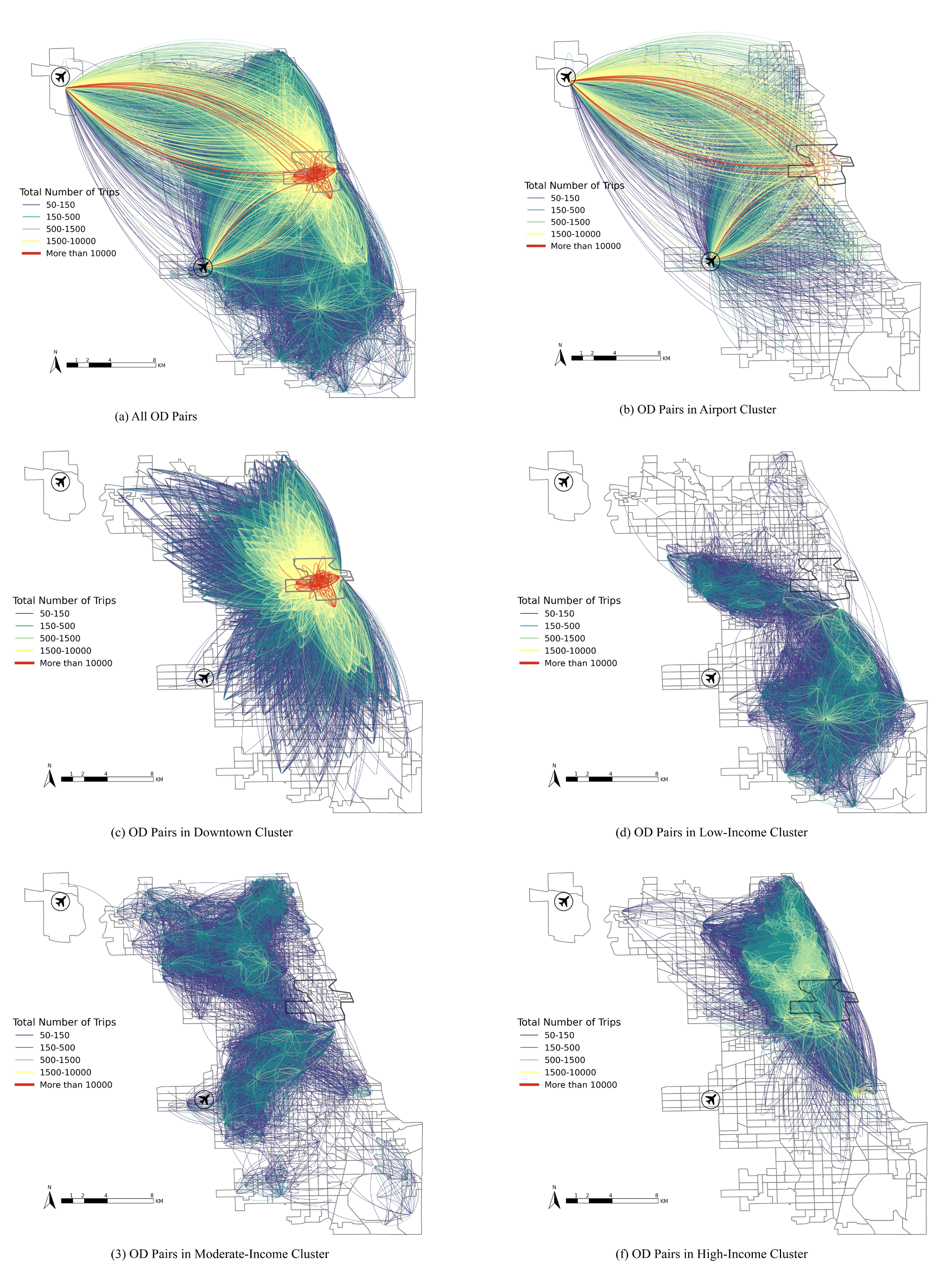}
  \caption{Spatial Patterns of Five Clusters and All OD Pairs }\label{fig:OD pairs 1114}
\end{figure}

\subsection{Cluster Profiles}
\label{cluster profile}
Since it is difficult to explore meaningful spatial patterns of five clusters with 67,498 OD pairs in a single map, we adopt the following visualization strategies. In order to figure out the cluster-specific spatial variations of ridesourcing demand across the whole city, we link the origins and destinations of all OD pairs with the arcs that curved toward right in each cluster. The brighter and wider the OD flow line is, the more trips this OD pair has. We also highlight the downtown census tracts with black lines of their boundaries and add a ``plane'' icon on each airport tract. These visualization strategies allow us to have a more clear understanding of the spatial patterns and characteristics of each cluster. But there are two limitations that we cannot visualize OD pairs whose origins and destinations are the same tracts. \autoref{fig:OD pairs 1114} presents the visualization results.

\autoref{tab:OD_Pair_Characteristics} presents a brief profile of each cluster, including the average ridesourcing demand and the share (\%) of OD pairs per cluster. Accordingly, OD pairs were divided into five clusters: the largest cluster (Cluster 5) contains around 33.2\% OD pairs, while the smallest one (Cluster 1) accounts for 3\% approximately. Spatial patterns of five clusters clearly exhibit how the OD pairs with large ridesourcing demand distribute. Also, \autoref{fig:OD pairs 1114}(a) shows that most OD pairs with more than 10,000 trips are related to downtown and two airports. 

Cluster 1 is the smallest group containing all airport-related OD pairs, thus is named as \textbf{\textit{Airport Cluster}}. This cluster has obviously unique characteristics from the other four clusters as shown in \autoref{tab:Descriptive Statistics}. Notably, the values of transit-related variables are the lowest among the five clusters, indicating that the destinations and origins of airport-related trips in Chicago are not well served by public transit. This finding directly echoes with the results from \citep{soria2020k}, where they found that (1) origins and destinations of airport trips usually suffer poor transit connectivity and low taxi adoption rate; and (2) the estimated transit travel time for airport-related trips were more than 70\% longer than the others. Thus, the City of Chicago may consider plans to improve the transit coverage for airport trips, such as increasing the infrastructure construction. Nevertheless, it should be noted that the complementary effect is not always the indicator to plan more infrastructures, since the travel demand may not satisfy the daily operational cost of the transits \citep{kong2020does}. Moreover, we find that this cluster has only 1,981 OD pairs (3\%) while covering 626 census tracts as trip origins and 569 census tracts as trip destinations. However, it has the maximum average number of trips among five clusters. This finding suggests there is huge ridesourcing demand for airport rides.

All OD pairs in Cluster 2 are related to the downtown area in Chicago; thus this cluster is named as \textbf{\textit{Downtown Cluster}}. The number of OD pairs in Cluster 2 is 18,868 (27.95\%), including 701 census tracts as trip origins and 671 census tracts as trip destinations. There are several major findings from this cluster. Noticeably, most high-demand OD pairs are within downtown areas. This is explainable because: (1) downtown areas in Chicago have high population, employment and road network density, these characteristics collectively have a positive impact on ridesourcing demand; and (2) downtown areas are more likely to generate social, recreational and commuting activities, which play key roles in attracting riders to use ridesourcing services \citep{yan2020using,ghaffar2020modeling,marquet2020spatial}. Secondly, we find that even though the transit supply is the highest, the average ridesourcing demand in \textit{Downtown Cluster} is still considerable ($n$ = 646, the second highest). However, we cannot clearly distinguish the substitutional or complementary effects. Previous research has suggested that the substitutions aggregate in the city center \citep{kong2020does}. Also, the high ridesourcing usage might be attributed to the complementary effect, probably because the ridesourcing services can serve as feeders for public transit and complement it at late night and weekends \citep{jin2018ridesourcing}. Therefore, further investigations are required to identify the relationships between ridesourcing services and public transit for the downtown trips.

OD pairs in Cluster 3 have the second lowest average ridersourcing demand (135.87), including 401 census tracts as trip origins while 417 census tracts as trip destinations. Most of the OD pairs with high ridesourcing demand in Cluster 3 are concentrated in areas with low household income and distributed at central west, southeastern and far south areas in Chicago. These areas have relatively high crime density, low education and income levels, along with poor employment supply. Hence, we termed Cluster 3 as \textbf{\textit{Low-Income Cluster}}. Extensive studies found that ridesourcing demand is positively correlated with household income, education attainment and some built-environment variables (e.g., population and employment density) \citep{brown2018ridehail, yu2019exploring,ghaffar2020modeling, lavieri2018model}. Accordingly, the values of these variables in this cluster are relatively low, which is likely the reason why the mean ridesourcing demand here is the second lowest. This finding may inform transportation planners to develop incentive strategies to promote the adoption and affordability of ridesourcing services across low-income communities.

Compared with the characteristics in Cluster 3, we name Cluster 4 as \textbf{\textit{Moderate-Income Cluster}}. OD pairs in Cluster 4 primarily distribute at far north, northwestern and southwestern areas in Chicago. These areas consist of most of the Hispanic communities in Chicago, which is consistent with the finding (from \autoref{tab:Descriptive Statistics}) that origins and destinations in Cluster 4 have the highest proportion of Hispanic people. An interesting finding is that OD pairs in this cluster has the lowest mean demand for ridesourcing services. This finding further reinforces the previous finding that the ridesouricng adoption rate is lower in Hispanic communities \citep{marquet2020spatial,alemi2018influences}. One possible explanation is that census tracts in this cluster have the highest rate of car ownership (see the variable \textit{Pctcarown}). Intuitively, this is consistent with our expectation that households with more than one vehicle are more likely to fulfill the travel needs by themselves. Theoretically, this finding echoes one of the findings in \citep{sikder2019uses}, where they suggested that families with ``sufficient vehicles'' were less likely to use ridesourcing services. Moreover, census tracts in Cluster 4 have the second lowest rate of workers who take transit to work and second lowest level of transit supply. Previous research about ridesourcing demand modeling has identified a positive relationship between transit supply and ridesourcing demand \citep{brown2018ridehail, yu2019exploring}. Thus, it is reasonable that the mean demand for ridesourcing services of Cluster 4 is still the lowest among the five clusters.

In Cluster 5, a significant proportion of the OD pairs with high ridesourcing demand are concentrated in northern regions surrounding the downtown like ``Lincoln Park'' and ``Lake View'', while a minor proportion is distributed in the south of downtown, for example, the ``Near South Side''. Communities in these regions usually have high-level household income, high population density, and high transit accessibility. Origins and destinations in this cluster have the highest rate of young generations. Thus, Cluster 5 is named as \textbf{\textit{High-Income Cluster}}. Remarkably, there are 22,409 OD pairs (33.2\%) in Cluster 5, encompassing 442 census tracts as trip destinations and 448 census tracts as trip origins. However, its average number of trips (195.74) is more than that of \textit{Low-Income Cluster} and \textit{Moderate-Income Cluster}. These facts indicate that ridesourcing services are more popular and acceptable among wealthy and/or young people than the poor and/or senior people, which is consistent with the findings in \citep{clewlow2017disruptive, yu2019exploring, gerte2018there}.

\subsection{Model Performance}

\autoref{tab:comparison of the benchmarks} introduces the performance of different models when selected as the benchmark. Due to the inflexible model structure and ignorance of spatial heterogeneity, the traditional statistical models have rather lower prediction performance. As the benchmark model becomes more flexible, the predictive error significantly decreases. It confirms that the flexible machine learning models can effectively capture the underlying complex relationships behind the data and hence produce better prediction accuracy. In addition, compared with the best benchmark model (i.e., GBDT), the MAE and RMSE of CEM are still improved by 14.36\% and 6.78\%, respectively, which suggests the superiority of the CEM. A more transparent and flexible model structure allows CEM to cope with spatial heterogeneity and boost the predictive performance.

Setting GBDT as the benchmark model, we also present the accuracy improvement rate, i.e., error reduction rate, of MAE and RMSE for each cluster, as shown in \autoref{tab:Model Performance}. Together with \autoref{tab:comparison of the benchmarks}, we conclude that the proposed CEM has better prediction performance than all benchmark models. The cluster-specific submodels could simultaneously capture the nonlinear relationships and address the local heterogeneous effects within the cluster, hence reducing the prediction error. Nevertheless, the benchmark considers all OD pairs as a whole under the global context. They cannot accommodate enough local spatial variations, and therefore having poor demand estimations.

From \autoref{tab:Model Performance}, experimental results show that compared with the benchmark model, CEM has increased the predictive performance in Cluster 1 (i.e., \textit{Airport Cluster}) and Cluster 2 (i.e., \textit{Downtown Cluster}). Their cluster-wise accuracy improvement rate for MAE and RMSE are 7.45\%, 7.01\% and 6.65\%, 5.78\%, respectively. However, the improvement is modest compared with the other clusters, possibly due to three reasons. First, the standard deviation of data in \textit{Airport Cluster} and \textit{Downtown Cluster} is extensively higher than that of the other clusters, which may make cluster-specific variations less tractable. Second, characteristics of two airports and downtown areas are evidently different compared with other areas, indicating there exists a strong heterogeneous effect at the spatial scale. Third, we ignore some endogenous factors, which plays key role in impacting ridesourcing demand \citep{wang2019ridesourcing}. Specifically, in the City of Chicago, riders should pay a \$5 surcharge when using ridesourcing services for airport trips. Also for downtown trips, the TNCs always impose a surge pricing strategy to control the supply (driver availability) and travel demand. In most time, the TNCs implemented surge pricing strategies in downtown areas in Chicago, especially during peak hours. A previous study found that the percentage of time that Uber adopted surge pricing in Chicago is 24.7\% \citep{cohen2016using}. These static and dynamic pricing strategies play an important role in shaping both supply and demand \citep{wang2019ridesourcing}. Missing these endogenous factors such as trip surcharge and surge pricing strategies in modeling may limit the prediction accuracy. In addition, CEM delivers marginal better results in \textit{Airport Cluster} probably because this cluster has the smallest sample size (1,981 OD pairs) compared to other clusters. For \textit{Airport Cluster} submodel, the sample size might be relatively sufficient, whereas for the benchmark model, there will be more difficulties to fully capture the cluster-specific variations.

Compared to the best benchmark model, CEM yields substantially superior predictive results on Cluster 3 (i.e., \textit{Low-Income Cluster}), Cluster 4 (i.e., \textit{Moderate-Income Cluster}) and Cluster 5 (i.e., \textit{High-Income Cluster}), with an average accuracy improvement rate for MAE and RMSE of 24.36\% and 17.30\%, respectively. This is reasonable because (1) the standard deviation of variables in this cluster are the lowest, suggesting the data are less dispersed; and (2) these clusters are generated from the data-driven clustering, which makes their patterns unique and easy to capture. A cluster-level machine-learning model can more efficiently address the location-specific nature of ridesourcing-trip data and improves the bias-variance trade-off, which is directly in line with the findings in \citet{trivedi2015utility}. Therefore, CEM exhibits superior predictive performance.

\begin{table}[!ht]
\centering
\caption{Performance Comparison of the Benchmarks}
\label{tab:comparison of the benchmarks}
\resizebox{0.35\textwidth}{!}{%
\begin{tabular}{@{}lll@{}}
\toprule
Model & MAE & RMSE \\ \midrule
\textit{Benchmark} &  &  \\
DT & 172.04 & 841.92 \\
RF & 106.59 & 364.07 \\
GBDT & 110.50 & 318.15 \\
XGBoost & 115.50 & 333.41 \\
SVM & 197.81 & 860.67 \\
ANN & 213.76 & 610.07 \\
LR & 328.51 & 858.22 \\
Log-LR & 182.31 & 941.71 \\
Poisson & 201.42 & 720.29 \\ \midrule
\textit{Proposed Model} &  &  \\
CEM & 94.64 & 296.57 \\ \bottomrule
\end{tabular}%
}
\end{table}

\begin{table}[!ht]
\centering
\caption{Performance of CEM Compared with the Best Benchmark}
\label{tab:Model Performance}
\resizebox{0.9\textwidth}{!}{%
\begin{tabular}{@{}ccccccc@{}}
\toprule
\multirow{2}{*}{Testing Set} & \multicolumn{2}{c}{CEM} & \multicolumn{2}{c}{The Best Benchmark} & \multicolumn{2}{l}{Accuracy Improvement Rate} \\ \cmidrule(l){2-7} 
 & MAE & RMSE & MAE & RMSE & MAE & RMSE \\ \midrule
Cluster 1 & 292.38 & 578.93 & 315.92 & 620.15 & 7.45\% & 6.65\% \\
Cluster 2 & 180.56 & 503.10 & 194.17 & 533.95 & 7.01\% & 5.78\% \\
Cluster 3 & 46.70 & 92.75 & 61.29 & 109.14 & 23.81\% & 15.01\% \\
Cluster 4 & 44.43 & 77.47 & 59.53 & 102.36 & 25.36\% & 24.32\% \\
Cluster 5 & 57.87 & 119.42 & 76.05 & 136.62 & 23.91\% & 12.58\% \\ \bottomrule
\end{tabular}%
}
\end{table}

\section{Discussion and Conclusion}
\label{S:Conclusion}

This paper presents a study of ridesourcing demand prediction in the City of Chicago. Using the publicly released ridesourcing trip data in Chicago and merging census-tract-level socioeconomic and demographic, built-environment and transit-supply data, this study proposes a Clustering-aided Ensemble Method (CEM) to predict the ridesourcing demand of each OD (tract-to-tract) pairs. Combining the domain knowledge and previous studies, we first separate OD pairs containing airport and downtown tracts as origins or destinations as two special clusters. Then, we apply clustering to divide remaining OD pairs into three clusters. Next, we adopt several machine-learning models to build the cluster-specific models. After selecting each cluster-specific best-performing submodel, we aggregate them together to form the CEM. We then compare the predictive capabilities of CEM with the benchmark model (built with the entire training set). Empirical results show that, with a more transparent and flexible (i.e., great flexibility in using any type of cluster-specific model) model structure, CEM achieved significantly better prediction performance than the benchmark.

Through cluster analysis, all OD pairs are divided into five clusters. Based on their characteristics and spatial distribution, we name these five clusters as \textit{Airport Cluster}, \textit{Downtown Cluster}, \textit{Low-Income Cluster}, \textit{Moderate-Income Cluster} and \textit{High-Income Cluster}. These findings not only enable transportation planners to improve the ridesourcing demand forecasting, but also provide a reference for ridesourcing providers to distinguish ridesourcing market segments.

It should be noted that the main purpose of this study is to introduce a new machine-learning model, CEM, and its application of predicting the ridesourcing demand. Some interesting insights and policy implications would be generated if we further delve into the determinants of each cluster-specific submodel. However, due to the limited space, we only focus on introducing a more accurate approach to predict the ridesourcing demand rather than investigating its determinants. In recent years, a large body of literature has adopted machine learning models to study how key factors can impact the travel behavior and extract various valuable insights from the results \citep[e.g.][]{ding2018applying, ding2019does, zhao2020prediction}. However, most of the previous studies (especially for machine-learning studies) modeled the travel behavior globally and seldom controlled the spatial heterogeneity. In reality, the impact of the associated factors is multi-faceted and the nature of the travel behavior has strong spatial heterogeneity. We may wonder how and to what extent do these factors vary across different spatial contexts. To bridge this gap, future studies could use the state-of-the-art machine learning interpretation methods (e.g., feature importance and partial dependence plot) together with our proposed framework to explore the travel behavior of different segmentations (e.g., user group, survey respondents) or of different spatial units (e.g., traffic analysis zones or census block groups). Overall, this study contributes to the current literature by offering a guidance for future behavioral research.

Some issues require further investigation. Due to the model complexity, the temporal attributes were not incorporated as variables to predict ridesourcing demand in this study. We recommend future studies to consider the impacts of the time of day on ridesourcing demand since the demand distributions might be distinct during peak and off-peak hours. Another limitation of our research is that we only used ridesourcing-trip data in Chicago as a case study, thus the results may not be directly transferable to other cities or travel modes.

\section*{Authorship Contribution Statement}
\textbf{Zhang}: Conceptualization, Data Curation, Methodology, Software, Formal Analysis, and Draft Preparation. \textbf{Zhao}: Conceptualization, Methodology, Formal Analysis, Draft Preparation, Supervision and Grant Acquisition.

\section*{Acknowledgment}
This research was partially supported by the U.S. Department of Transportation through the Southeastern Transportation Research, Innovation, Development and Education (STRIDE) Region 4 University Transportation Center (Grant No. 69A3551747104). We are grateful for the valuable comments and suggestions provided by Xiang `Jacob' Yan, Xinyu Liu and Yiming Xu.

\FloatBarrier

\section*{Appendix}
\label{Appendix}

\subsection*{Appendix.A}

To find the optimal values of all hyperparameters, we applied \textit{Grid Search} with a 5-fold CV on every tree-based model when training each cluster-specific subset and the entire dataset. We use MSE as the metric to evaluate each model's predictive performance. The detailed information of all hyperparameters is presented in \autoref{tab:Optimal Values of Hyperparameterse}.

\begin{table}[H]
\caption{Optimal Values of Hyperparameters}
\label{tab:Optimal Values of Hyperparameterse}
\resizebox{\textwidth}{!}{%
\begin{tabular}{@{}lllllll@{}}
\toprule
 & DT & RF & GBDT & XGBoost & SVM & NN \\ \midrule
\begin{tabular}[c]{@{}l@{}}Cluster 1\\ (Airport)\end{tabular} & \begin{tabular}[c]{@{}l@{}}\# of featues = 40, \\ min\_sample\_split = 10,\\ complex parameter = 0.001\end{tabular} & \begin{tabular}[c]{@{}l@{}}\# of trees = 200, \\ \# of features = 6\end{tabular} & \begin{tabular}[c]{@{}l@{}}\# of trees = 400, \\ \# of features = 6, \\ learning\_rate = 0.05, \\ depth = 5\end{tabular} & \begin{tabular}[c]{@{}l@{}}\# of trees = 400, \\ \# of features = 8, \\ learning\_rate = 0.04, \\ depth = 5\end{tabular} & \begin{tabular}[c]{@{}l@{}}C = 10000, \\ kernal = 'rbf'\end{tabular} & \begin{tabular}[c]{@{}l@{}}hidden layer size = 1,\\ weight decay = 0.1,\\ number of neurons = 30,\\ learning\_rate = 0.1\end{tabular} \\
\begin{tabular}[c]{@{}l@{}}Cluster 2\\ (Downtown)\end{tabular} & \begin{tabular}[c]{@{}l@{}}\# of featues = 50, \\ min\_sample\_split = 10,\\ complex parameter = 0.019\end{tabular} & \begin{tabular}[c]{@{}l@{}}\# of trees = 400, \\ \# of features = 8\end{tabular} & \begin{tabular}[c]{@{}l@{}}\# of trees = 400, \\ \# of features = 8, \\ learning\_rate = 0.05, \\ depth = 5\end{tabular} & \begin{tabular}[c]{@{}l@{}}\# of trees = 400, \\ \# of features = 8, \\ learning\_rate = 0.05, \\ depth = 5\end{tabular} & \begin{tabular}[c]{@{}l@{}}C = 10000, \\ kernal = 'rbf'\end{tabular} & \begin{tabular}[c]{@{}l@{}}hidden layer size = 1,\\ weight decay = 0.0001,\\ number of neurons = 30,\\ learning\_rate = 0.0005\end{tabular} \\
\begin{tabular}[c]{@{}l@{}}Cluster 3\\ (Low Income)\end{tabular} & \begin{tabular}[c]{@{}l@{}}\# of featues = 10, \\ min\_sample\_split = 26,\\ complex parameter = 0.019\end{tabular} & \begin{tabular}[c]{@{}l@{}}\# of trees = 400, \\ \# of features = 6\end{tabular} & \begin{tabular}[c]{@{}l@{}}\# of trees = 400, \\ \# of features = 6, \\ learning\_rate = 0.05, \\ depth = 5\end{tabular} & \begin{tabular}[c]{@{}l@{}}\# of trees = 400, \\ \# of features = 8, \\ learning\_rate = 0.05, \\ depth = 5\end{tabular} & \begin{tabular}[c]{@{}l@{}}C = 10000, \\ kernal = 'rbf'\end{tabular} & \begin{tabular}[c]{@{}l@{}}hidden layer size = 1,\\ weight decay = 0.0001,\\ number of neurons = 50,\\ learning\_rate = 0.0001\end{tabular} \\
\begin{tabular}[c]{@{}l@{}}Cluster 4\\ (Moderate Income)\end{tabular} & \begin{tabular}[c]{@{}l@{}}\# of featues = 10, \\ min\_sample\_split = 22,\\ complex parameter = 0.016\end{tabular} & \begin{tabular}[c]{@{}l@{}}\# of trees = 300,\\  \# of features = 6\end{tabular} & \begin{tabular}[c]{@{}l@{}}\# of trees = 400, \\ \# of features = 8, \\ learning\_rate = 0.05, \\ depth = 5\end{tabular} & \begin{tabular}[c]{@{}l@{}}\# of trees = 400, \\ \# of features = 8, \\ learning\_rate = 0.05, \\ depth = 5\end{tabular} & \begin{tabular}[c]{@{}l@{}}C = 10000, \\ kernal = 'rbf'\end{tabular} & \begin{tabular}[c]{@{}l@{}}hidden layer size = 1,\\ weight decay = 0.5,\\ number of neurons = 30,\\ learning\_rate = 0.0001\end{tabular} \\
\begin{tabular}[c]{@{}l@{}}Cluster 5\\ (High Income)\end{tabular} & \begin{tabular}[c]{@{}l@{}}\# of featues = 20, \\ min\_sample\_split = 20,\\ complex parameter = 0.019\end{tabular} & \begin{tabular}[c]{@{}l@{}}\# of trees = 200, \\ \# of features = 8\end{tabular} & \begin{tabular}[c]{@{}l@{}}\# of trees = 400, \\ \# of features = 6, \\ learning\_rate = 0.05, \\ depth = 5\end{tabular} & \begin{tabular}[c]{@{}l@{}}\# of trees = 400, \\ \# of features = 8, \\ learning\_rate = 0.05, \\ depth = 5\end{tabular} & \begin{tabular}[c]{@{}l@{}}C = 10000, \\ kernal = 'rbf'\end{tabular} & \begin{tabular}[c]{@{}l@{}}hidden layer size = 1,\\ weight decay = 0.1,\\ number of neurons = 30,\\ learning\_rate = 0.005\end{tabular} \\
All OD Pairs & \begin{tabular}[c]{@{}l@{}}\# of featues = 20, \\ min\_sample\_split = 28,\\ complex parameter = 0.004\end{tabular} & \begin{tabular}[c]{@{}l@{}}\# of trees = 300, \\ \# of features = 8\end{tabular} & \begin{tabular}[c]{@{}l@{}}\# of trees = 400, \\ \# of features = 8, \\ learning\_rate = 0.04, \\ depth = 5\end{tabular} & \begin{tabular}[c]{@{}l@{}}\# of trees = 400, \\ \# of features = 8, \\ learning\_rate = 0.05, \\ depth = 5\end{tabular} & \begin{tabular}[c]{@{}l@{}}C = 10000, \\ kernal = 'rbf'\end{tabular} & \begin{tabular}[c]{@{}l@{}}hidden layer size = 1,\\ weight decay = 0.001,\\ number of neurons = 20,\\ learning\_rate = 0.005\end{tabular} \\ \bottomrule
\end{tabular}%
}
\end{table}

\subsection*{Appendix.B}

We computed the DBI under different iterations. We set the number of clusters to range from 2 to 7 and perform 100 iterations for each scenario. According to \autoref{tab:Clustering Model Comparison}, \textit{K-Means} with 3 clusters is finally chosen for the data-driven clustering as it achieves the lowest mean value of DBI (2.09).

\begin{table}[!ht]
\centering
\caption{Clustering Model Comparison through DBI}
\label{tab:Clustering Model Comparison}
\resizebox{0.75\textwidth}{!}{%
\begin{tabular}{@{}lllllll@{}}
\toprule
Number of Clusters     & 2     & 3     & 4     & 5     & 6     & 7     \\ \midrule
DBI of KMeans                 & 2.12 & \textbf{2.09} & 2.44 & 2.46 & 2.50 & 2.43 \\
DBI of Gaussian-Mixture Model & 2.35 & 4.24 & 3.39 & 3.19 & 3.10 & 2.98 \\ \bottomrule
\end{tabular}%
}
\end{table}

\subsection*{Appendix.C}

The descriptive statistics for each cluster is presented in \autoref{tab:Descriptive Statistics}. To make it transparent for readers, we present them with the pre-normalized data.

\begin{table}[!h]
\centering
\caption{Descriptive Statistics}
\label{tab:Descriptive Statistics}
\begin{threeparttable}

\resizebox{\textwidth}{!}{%
\begin{tabular}{@{}lllllllllll@{}}
\toprule
\multirow{2}{*}{Variable} & \multicolumn{2}{l}{Cluster 1} & \multicolumn{2}{l}{Cluster 2} & \multicolumn{2}{l}{Cluster 3} & \multicolumn{2}{l}{Cluster 4} & \multicolumn{2}{l}{Cluster 5} \\ \cmidrule(l){2-11} 
 & Mean & SD & Mean & SD & Mean & SD & Mean & SD & Mean & SD \\ \midrule
Total\_number\_trips & 819.30 & 2489.61 & 646.00 & 2211.85 & 135.87 & 159.14 & 128.50 & 117.77 & 195.74 & 286.74 \\
Fare\_sd & 6.33 & 2.09 & 3.80 & 1.23 & 2.53 & 0.83 & 2.29 & 0.76 & 2.48 & 1.10 \\
Fare\_median & 22.16 & 7.53 & 11.50 & 3.68 & 7.39 & 2.82 & 6.80 & 2.51 & 8.12 & 3.24 \\
Miles\_sd & 1.32 & 0.90 & 0.92 & 0.44 & 1.03 & 0.38 & 0.86 & 0.35 & 0.69 & 0.35 \\
Miles\_median & 14.41 & 6.08 & 6.46 & 3.10 & 3.50 & 2.32 & 2.93 & 1.86 & 3.63 & 2.38 \\
Seconds\_sd & 593.81 & 202.54 & 402.26 & 147.59 & 275.12 & 96.66 & 263.81 & 104.33 & 253.91 & 121.13 \\
Seconds\_median & 1852.37 & 552.19 & 1223.12 & 424.10 & 713.52 & 348.94 & 700.03 & 350.67 & 841.07 & 398.44 \\
Pcttransit\_Ori & 0.17 & 0.18 & 0.31 & 0.12 & 0.32 & 0.10 & 0.24 & 0.09 & 0.41 & 0.11 \\
Pctmidinc\_Ori & 0.14 & 0.14 & 0.26 & 0.07 & 0.22 & 0.08 & 0.29 & 0.06 & 0.27 & 0.07 \\
Pctmale\_Ori & 0.26 & 0.25 & 0.49 & 0.05 & 0.45 & 0.05 & 0.50 & 0.04 & 0.51 & 0.04 \\
Pctsinfam\_Ori & 0.14 & 0.22 & 0.13 & 0.16 & 0.31 & 0.26 & 0.32 & 0.21 & 0.16 & 0.11 \\
Pctmodinc\_Ori & 0.11 & 0.12 & 0.14 & 0.09 & 0.23 & 0.07 & 0.25 & 0.08 & 0.16 & 0.07 \\
Pctyoung\_Ori & 0.25 & 0.25 & 0.56 & 0.14 & 0.39 & 0.09 & 0.44 & 0.09 & 0.59 & 0.11 \\
Pctwhite\_Ori & 0.28 & 0.35 & 0.63 & 0.25 & 0.11 & 0.17 & 0.59 & 0.21 & 0.75 & 0.17 \\
Pcthisp\_Ori & 0.13 & 0.23 & 0.15 & 0.20 & 0.07 & 0.14 & 0.49 & 0.28 & 0.18 & 0.16 \\
Pctcarown\_Ori & 0.39 & 0.38 & 0.66 & 0.15 & 0.64 & 0.13 & 0.81 & 0.09 & 0.73 & 0.13 \\
Pctrentocc\_Ori & 0.31 & 0.32 & 0.60 & 0.15 & 0.66 & 0.19 & 0.53 & 0.16 & 0.61 & 0.13 \\
Pctasian\_Ori & 0.04 & 0.08 & 0.11 & 0.10 & 0.04 & 0.11 & 0.07 & 0.10 & 0.08 & 0.07 \\
Pctlowinc\_Ori & 0.14 & 0.17 & 0.21 & 0.13 & 0.43 & 0.13 & 0.24 & 0.10 & 0.18 & 0.12 \\
CrimeDen\_Ori & 68.81 & 125.36 & 243.46 & 263.51 & 249.36 & 188.93 & 105.40 & 88.76 & 116.43 & 89.46 \\
PctWacWorker54\_Ori & 0.78 & 0.05 & 0.80 & 0.05 & 0.77 & 0.07 & 0.77 & 0.05 & 0.81 & 0.06 \\
PctWacLowMidWage\_Ori & 0.56 & 0.19 & 0.60 & 0.18 & 0.71 & 0.17 & 0.72 & 0.13 & 0.70 & 0.13 \\
PctWacBachelor\_Ori & 0.21 & 0.05 & 0.23 & 0.07 & 0.18 & 0.07 & 0.19 & 0.06 & 0.21 & 0.06 \\
Commuters\_HW & 5.98 & 11.70 & 17.00 & 60.31 & 1.88 & 6.24 & 3.49 & 6.19 & 2.71 & 8.30 \\
Commuters\_WH & 5.53 & 11.60 & 16.59 & 60.39 & 1.79 & 6.23 & 3.25 & 6.09 & 2.69 & 8.28 \\
Popden\_Ori & 11464.10 & 16170.07 & 28578.29 & 19884.29 & 13090.77 & 7009.51 & 18302.63 & 8956.21 & 25684.23 & 14374.34 \\
IntertDen\_Ori & 63.94 & 76.88 & 188.59 & 157.04 & 90.27 & 53.82 & 93.29 & 50.41 & 129.86 & 87.36 \\
EmpDen\_Ori & 7355.76 & 26934.16 & 67875.80 & 132529.45 & 2826.36 & 5653.59 & 3972.63 & 4548.65 & 8492.04 & 9611.38 \\
EmpRetailDen\_Ori & 517.42 & 2531.34 & 4444.46 & 9791.31 & 290.74 & 546.15 & 529.14 & 671.27 & 1051.17 & 1609.80 \\
EmpRetailDen\_Des & 503.61 & 2530.12 & 4694.73 & 10044.96 & 294.85 & 546.42 & 524.74 & 670.62 & 1041.48 & 1599.98 \\
Walkscore\_Ori & 55.87 & 29.16 & 85.37 & 15.60 & 67.57 & 15.94 & 75.87 & 14.23 & 85.68 & 13.88 \\
Walkscore\_Des & 53.09 & 29.85 & 85.83 & 15.49 & 67.38 & 16.21 & 75.19 & 14.85 & 85.56 & 14.01 \\
RdNetwkDen\_Ori & 15.36 & 10.74 & 31.01 & 11.00 & 23.37 & 5.48 & 22.76 & 5.11 & 25.99 & 7.32 \\
SerHourBusRoutes\_Ori & 760.58 & 613.90 & 1943.09 & 1261.49 & 952.84 & 374.20 & 837.61 & 378.41 & 1200.08 & 470.25 \\
SerHourRailRoutes\_Ori & 269.03 & 259.00 & 693.50 & 596.92 & 176.73 & 241.85 & 130.74 & 203.86 & 397.24 & 308.51 \\
PctBusBuf\_Ori & 0.63 & 0.39 & 0.93 & 0.15 & 0.91 & 0.16 & 0.94 & 0.12 & 0.96 & 0.12 \\
PctRailBuf\_Ori & 0.12 & 0.22 & 0.36 & 0.35 & 0.12 & 0.21 & 0.08 & 0.18 & 0.31 & 0.33 \\
BusStopDen\_Ori & 37.44 & 33.79 & 80.73 & 47.46 & 56.54 & 24.39 & 53.22 & 21.33 & 66.21 & 29.30 \\
RailStationDen\_Ori & 0.75 & 2.04 & 2.94 & 5.00 & 0.70 & 1.44 & 0.43 & 1.46 & 1.95 & 3.44 \\ \bottomrule
\end{tabular}%
}

\parbox[t]{0.98\textwidth}{\vskip3pt{\footnotesize$^a$Cluster 1: Airport Cluster; Cluster 2: Downtown Cluster; Cluster 3: Low-Income Cluster; Cluster 4: Moderate-Income Cluster; Cluster 5: High-Income Cluster.}}

\end{threeparttable}
\end{table}

\FloatBarrier
\newpage





\bibliographystyle{elsarticle-harv}
\biboptions{semicolon,round,sort,authoryear}
\bibliography{sample.bib}







\end{document}